\documentclass[letterpaper]{article} 
\usepackage{aaai2027}  
\usepackage[hyphens]{url}  
\usepackage{graphicx} 
\usepackage{natbib}  
\usepackage{caption} 
\usepackage{algorithm}
\usepackage{algorithmic}
\usepackage{booktabs}
\usepackage{tabularx}
\usepackage{newfloat}
\usepackage{listings}
\DeclareCaptionStyle{ruled}{labelfont=normalfont,labelsep=colon,strut=off} 
\floatstyle{ruled}
\newfloat{listing}{tb}{lst}{}
\floatname{listing}{Listing}

\usepackage{booktabs}

\usepackage{amsmath}
\usepackage{amssymb}   %
\usepackage{booktabs}  
\usepackage{multirow}  
\usepackage{amssymb}   
\title{SceneReGen: Generative Reconstruction of 3D Scenes from a Single Image}

\author{
    Zefan Tian\textsuperscript{\rm 1}\equalcontrib,
    Yuteng Ye\textsuperscript{\rm 1}\equalcontrib,
    Yiheng Zhang\textsuperscript{\rm 2},
    Yuhang Yang\textsuperscript{\rm 2},
    Xueqiang Lv\textsuperscript{\rm 2},
    Shizhou Zhang\textsuperscript{\rm 2},\\
    Le Liu\textsuperscript{\rm 2}\textsuperscript{†},  
    Di Xu\textsuperscript{\rm 1}\thanks{Corresponding author:  xudi21@huawei.com}   
}

\affiliations{
    \textsuperscript{\rm 1}Huawei\\
    \textsuperscript{\rm 2}Northwestern Polytechnical University
}

\begin{document}

\maketitle

\begin{abstract}
Single-image 3D scene reconstruction must complete partially observed objects and
place them coherently in a shared observation-aligned scene frame. Object-level
generative priors offer strong completion ability, but their centered,
scale-normalized outputs are typically expressed in an object frame, creating a
fundamental representation gap between object generation and scene reconstruction. We
introduce SceneReGen, a generative reconstruction framework that reinterprets scene reconstruction as the generation and assembly of complete object assets in a shared observation-aligned scene frame. SceneReGen addresses the generation–reconstruction gap through selective pose factorization: each object's observed
orientation is encoded directly in the generated mesh, while translation and scale are
estimated from instance-level and global scene evidence. Given a
scene image and instance masks, a geometry encoder extracts dense cues;
learnable shape queries condition a pretrained DiT-based 3D generator to produce
complete meshes in their observed orientations, while position queries fuse
object and scene features to assemble them in the shared frame. On the 3D-FUTURE
evaluation subset, SceneReGen achieves the best scene-level CD, scene-level
F-Score, and 3D bounding-box IoU among the evaluated methods, ties the best
object-level CD, and ranks second in object-level F-Score. Qualitative outputs
in autonomous-driving and embodied-AI scenes further illustrate the potential
of asset-centric reconstruction beyond indoor furniture.
\end{abstract}


\section{Introduction}

3D scene reconstruction from a single RGB image is fundamental for embodied AI, autonomous driving, and virtual content
creation. Given the image, the task requires recovering a complete mesh for
each partially observed object and placing the meshes coherently in a shared
observation-aligned scene frame.
Recent feed-forward 3D reconstruction models, such as
DUSt3R~\cite{wang2024dust3r} and VGGT~\cite{wang2025vggt}, recover useful
geometric cues from one or more views. In a single view, however, self- and
inter-object occlusions leave much of the object geometry unobserved. Recent 3D
generative models provide strong priors over complete object
geometry~\cite{zhang20233dshape2vecset,zhao2023michelangelo,zhang2024clay,
zhao2025hunyuan3d,xiang2025structured}, making them promising for completing
these missing regions.

Applying object-level generative priors to object-centric 3D scene
reconstruction, however, exposes the central representation gap addressed in
this work:
generated geometry is typically centered, scale-normalized, and expressed in a
canonical object frame, whereas the target scene requires each complete mesh to
be oriented, translated, and scaled in a shared observation-aligned scene
frame. Reconstruction must therefore bridge object completion and
observation-aligned spatial placement rather than solve either in isolation.
Existing approaches broadly follow two routes.
Object-first pipelines complete each object separately and then recover its
object-to-scene transformation~\cite{ardelean2025gen3dsr,yao2025cast}. Even
recent variants that use object-local cues for rotation and scene-level cues
for translation and scale still represent all three as explicit transformation
outputs~\cite{shi2026scenemaker}. These designs preserve dedicated object
priors, but keep object-to-scene placement separate from geometry generation.
With partial observations, errors in completed geometry or its correspondence
to visible evidence can then carry into placement. Joint
scene-modeling approaches instead use cross-object context to predict object
geometry and spatial arrangement within a shared
framework~\cite{huang2025midi,meng2026scenegen}. This brings scene context into
completion, but binds detailed geometry to the full placement problem, making
geometry and placement errors harder to separate. These complementary risks

\begin{figure*}[t]
    \centering
    \includegraphics[width=0.98\textwidth]{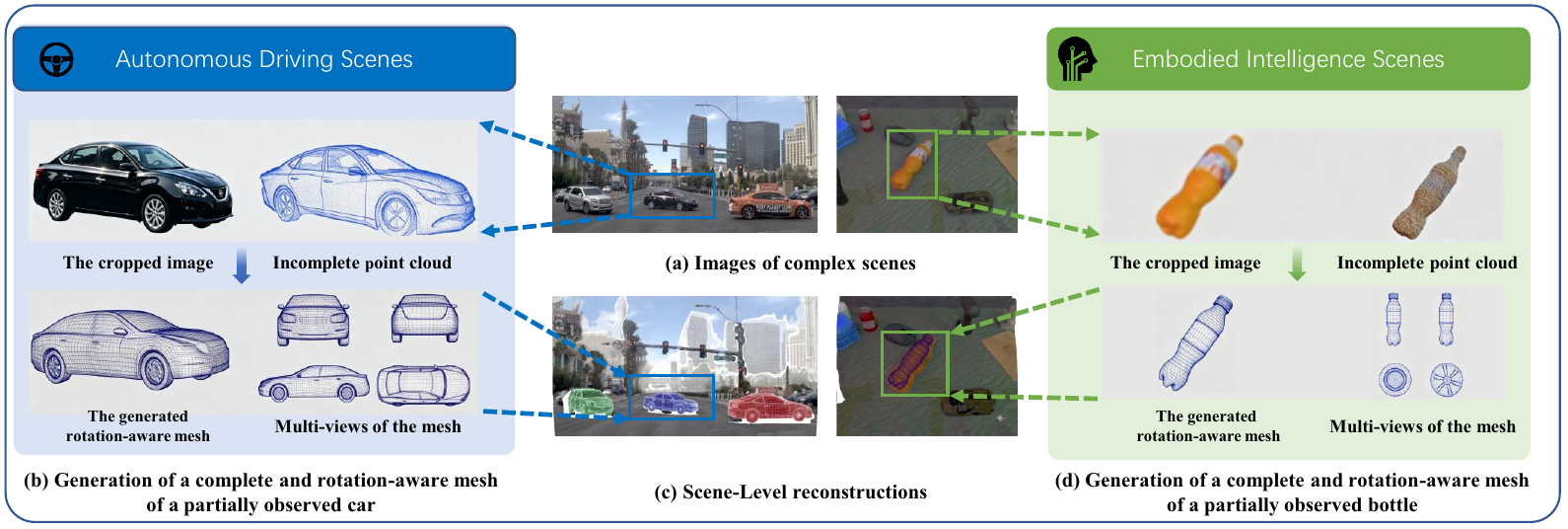}
    \caption{Qualitative SceneReGen outputs in autonomous-driving and embodied-AI
    scenarios. Given a scene image (a), SceneReGen reconstructs
    complete object meshes in their observed orientations (b, d) and estimates
    translation and scale to assemble them in a shared observation-aligned scene
    frame (c).}
    \label{fig:teaser}
\end{figure*}

We therefore ask a more targeted question: which pose components should be
coupled to object generation, and which should be estimated from scene evidence?
Our hypothesis is that orientation is more tightly coupled to image-conditioned
object generation. Completing a partially observed object requires mapping its
visible structures into a complete mesh in the target orientation; a post-hoc
rigid rotation can reorient that mesh, but does not by itself resolve
inconsistencies introduced during completion. Translation and scale instead
determine where the completed object lies and how large it is in the
shared observation-aligned scene frame. The 2D object extent, image-derived
depth and local geometry, and cross-object context provide complementary cues
for estimating these variables, although single-view ambiguity remains. This
motivates a selective pose factorization: orientation is represented in the
generated vertices, while translation and scale remain explicit placement
variables estimated from scene evidence.

Based on this design, we introduce \textbf{SceneReGen}, a framework that turns
rotation-aware object generation into single-image 3D scene reconstruction
through selective pose factorization. For each masked object instance, a
pretrained geometry encoder extracts dense image-derived geometric cues.
Learnable shape queries aggregate these cues into compact conditioning tokens
for a pretrained object-level 3D generator, which we fine-tune to produce a
complete mesh in the observed orientation while keeping it centered and
scale-normalized. Scene-aware position queries fuse instance-level and global
scene features from the geometry encoder to estimate translation and scale,
which place each generated mesh in the shared observation-aligned scene frame.

We train and evaluate SceneReGen on
3D-FUTURE~\cite{fu20213d} (Figure~\ref{fig:experiment_results} and Table~\ref{tab:sota_compare}). Figure~\ref{fig:teaser} additionally presents actual
model outputs on autonomous-driving and embodied-AI scene images, illustrating
the potential scope of complete object meshes in their observed orientations
and shared-frame scene assembly beyond the indoor-furniture benchmark.

Our contributions are summarized as follows:
\begin{itemize}

\item We formulate an object-centric generative reconstruction framework
for single-image 3D scene reconstruction, where generated object assets serve
as reconstruction primitives for scene recovery.

\item We identify the representation gap between object generation and scene
reconstruction and propose selective pose factorization, incorporating
orientation into generated geometry while estimating translation and scale from
scene evidence.

\item We develop SceneReGen, a geometry-aware generative reconstruction
framework that achieves superior scene-level accuracy and competitive
object-level quality on 3D-FUTURE.

\end{itemize}

\section{Related Work}
\subsection{3D Scene Reconstruction}
Traditional multi-view stereo and structure-from-motion pipelines such as COLMAP~\cite{schoenberger2016colmap} and PMVS~\cite{furukawa2009accurate} rely on sparse feature matching, yet often produce incomplete geometry in textureless or occluded areas. Feed-forward networks like DUSt3R~\cite{wang2024dust3r} and VGGT~\cite{wang2025vggt} accelerate reconstruction via direct regression, but still struggle with geometric sparsity and missing content. To address these limitations, VGGT-$\Omega$~\cite{wang2026vggt} builds upon the versatile geometric encoder VGGT, which bridges 2D visual observations with 3D geometric cues by modeling cross-view topological and spatial relationships, and further introduces register attention to distill compact, high-level geometric features.

%

\subsection{3D Object Generation}

The rapid progress of generative models~\cite{ho2020denoising, song2020denoising, papamakarios2021normalizing,NEURIPS2024_e25198b6} has significantly advanced 3D object generation. Early approaches typically distilled 2D generative priors from pre-trained text-to-image diffusion models to produce high-fidelity 3D representations. More recently, large-scale frameworks~\cite{li2025triposg,zhao2023michelangelo,zhao2025hunyuan3d,xiang2025structured,zadaianchuk2026reconstructiongeneration3dmultiobject} have pushed the state of the art by conditioning on either text or image inputs, enabling versatile and realistic 3D asset generation. To enhance multi-view consistency and alleviate the ambiguity of single-view inputs, another line of work~\cite{liu2023zero, shi2024mvdream, shi2023zero123++, huang2025mv, voleti2024sv3d} employs pre-trained image or video diffusion models to generate consistent multi-view images, which are then fused into a 3D representation. Notably, these methods typically generate objects in a canonical coordinate space, without considering their global placement or orientation within a scene.

%

\subsection{Generative Reconstruction}

Generative reconstruction leverages powerful generative priors to enhance 3D reconstruction from limited observations, and recent efforts in this direction can be broadly categorized into two paradigms: object-level and scene-level generative reconstruction.

\paragraph{Object-level generative reconstruction.} Reconstruction-guided 3D generative approaches~\cite{chang2025reconviagen, lin2026mix3r,Huang_2025_CVPR} have been proposed to combine the strength of diffusion priors with robust reconstruction networks. For instance, ReconviGen and Mix3R employ strong reconstruction backbones~\cite{wang2025vggt, wang2025pi} to constrain the denoising process, achieving single-view 3D reconstructions with high-fidelity geometry and texture. Despite their impressive performance, these methods operate in a canonical coordinate space and assume known camera poses, making them ill-suited for pose-free inputs—a common scenario in real-world captures.

\paragraph{Scene-level generative reconstruction.} Recent scene-level generative reconstruction~\cite{siddiqui2026shaper,yao2025cast,liu2025agentic} incorporate 3D generative priors~\cite{zhao2025hunyuan3d,mo2023dit} to refine object details or jointly predict objects along with their spatial placements (rotation, translation, and scale), conditioned on reconstruction cues~\cite{wang2025vggt}. However, jointly optimizing all spatial parameters within a single framework inevitably introduces parameter coupling, which often leads to scene-level misalignments.

Notably, the above methods share a common limitation: generation is performed in a canonical space, while reconstruction is conducted in the world coordinate system. In contrast, we explicitly decouple rotation from translation and scale, which rotation is predicted by the generation branch, translation and scale are estimated by the reconstruction branch. 

\begin{figure*}[t]
    \centering
    \includegraphics[width=0.99\linewidth]{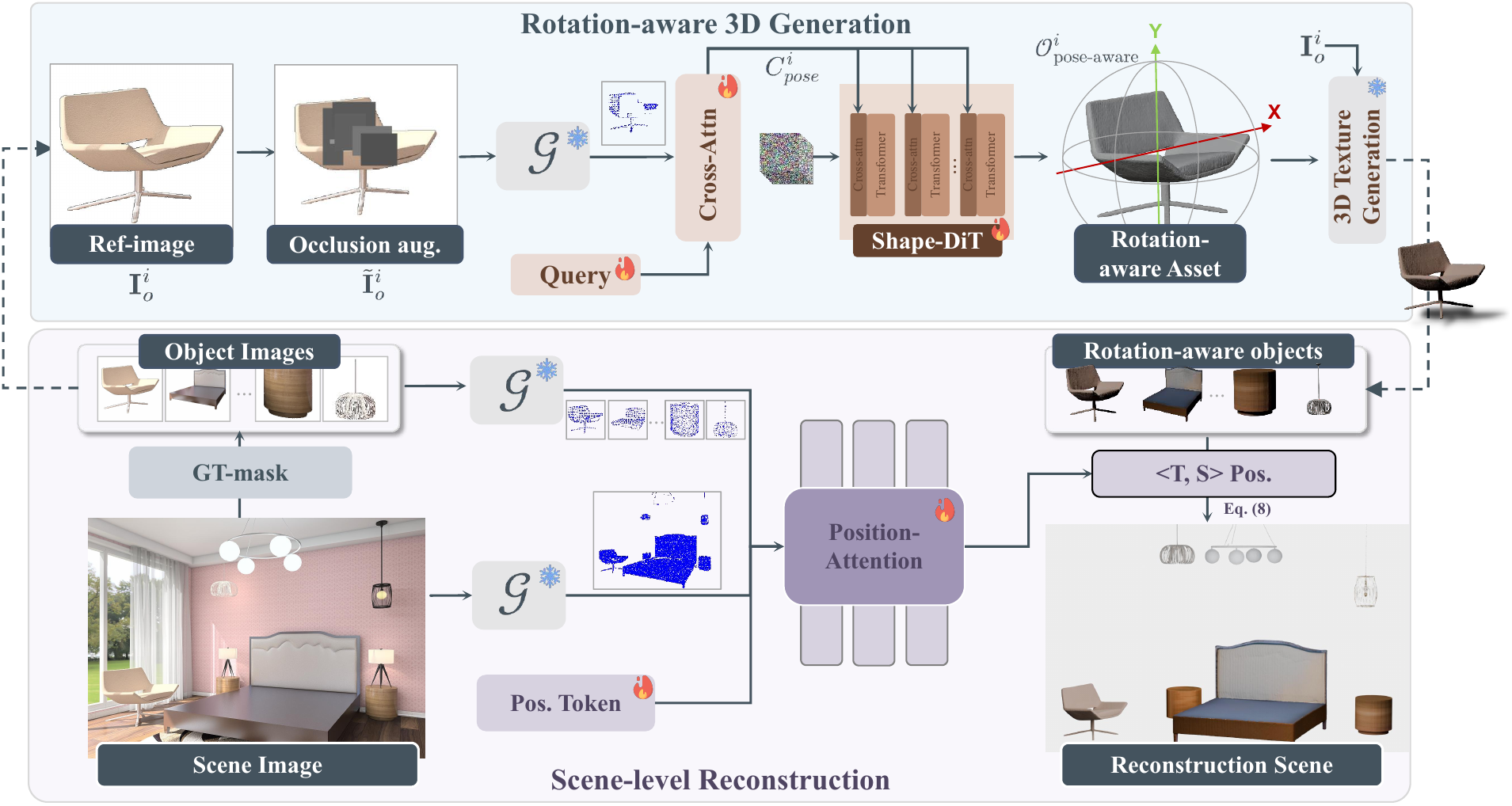}
    \caption{Overview of SceneReGen. Given a scene image and ground-truth instance
    masks, the rotation-aware generation branch uses object-level geometric cues
    to generate a complete mesh for each instance in its observed orientation,
    followed by texture generation. The scene-level reconstruction branch
    combines object- and scene-level cues with position queries to estimate
    translation and scale and assemble the objects in a shared
    observation-aligned scene frame.}
    \label{fig:pipeline}
\end{figure*}

\section{Method}
\label{sec:method}
Our goal is to reconstruct complete 3D object assets and their scene-level placements from a single image. SceneReGen combines rotation-aware object generation with scene-level position reconstruction. We formulate the problem in Sec.~\ref{Problem Definition}, describe object-wise generation in Sec.~\ref{subsec:pose_generation}, and present scene assembly through translation and scale estimation in Sec.~\ref{subsec:Position_reconstruction}.

\subsection{Problem Definition}
\label{Problem Definition}

Given a single scene image $\mathbf{I}$ containing $n$ object instances and the dataset-provided ground-truth instance masks $\mathcal{M} = \{\mathbf{M}_1, \dots, \mathbf{M}_n\}$, our framework reconstructs a complete 3D scene in a shared observation-aligned scene frame. Specifically, this frame is shared by all reconstructed objects and their estimated 3D positions, and inter-object spatial relationships are required to align with the object configuration observed in $\mathbf{I}$.

For each instance $i$, the generation branch produces a centered, scale-normalized mesh $\mathcal{O}_{\text{pose-aware}}^i$, in which the target orientation observed in $\mathbf{I}$ is encoded directly by the mesh vertices. The position branch then estimates a translation $\mathbf{t}_i \in \mathbb{R}^3$ and a positive isotropic scale $s_i \in \mathbb{R}_{>0}$. Applying the estimated scale and translation to the pose-aware meshes places all reconstructed objects in the shared observation-aligned scene frame, yielding the final 3D scene. Thus, rotation is represented by the generated geometry rather than regressed as a separate placement parameter.

\subsection{Rotation-Aware 3D Generation}
\label{subsec:pose_generation}

Using these instance masks, we isolate each object and its corresponding mask from the original scene image $\mathbf{I}$, obtaining object-specific RGB--mask pairs $\{(\mathbf{I}_o^i, \mathbf{M}_o^i)\}_{i=1}^n$. As illustrated in Figure~\ref{fig:pipeline}, the rotation-aware generation branch processes each pair independently to generate a complete pose-aware 3D mesh.

\paragraph{Occlusion Augmentation.} 
Since objects in real scenes are frequently occluded, directly using partial references often leads to incomplete generation. To enable the model to produce complete objects under such partial observations, we simulate occlusions during training by randomly placing \(5\) square masks that collectively cover \(20\%\) of the foreground pixels-a configuration chosen to reflect commonly encountered occlusion levels in real-world data, which encourage robust completion while preserving enough structural cues for learning-and filling the masked regions with random pixel values uniformly sampled from \([80, 180]\), a broad range chosen for occlusion diversity.

\paragraph{Geometry Condition.} After occlusion augmentation, we denote the processed object patch by $\tilde{\mathbf{I}}_o^i$ and feed it into a geometry encoder $\mathcal{G}$. In our implementation, $\mathcal{G}$ is instantiated with VGGT-$\Omega$~\cite{wang2026vggt}, a feed-forward reconstruction model built upon VGGT~\cite{wang2025vggt}. VGGT-$\Omega$ uses learned registers to aggregate scene information into a compact representation and introduces register attention for efficient information exchange through these registers. We select it because its geometry-aware representation provides global structural and spatial cues that are directly relevant to recovering an object's shape and observed orientation from a partial RGB observation.

For each object, we use VGGT-$\Omega$ to extract a sequence of geometry-aware features:
\begin{equation}
    \mathcal{F}_{obj}^i = \mathcal{G}(\tilde{\mathbf{I}}_o^i),
\end{equation}
where $\mathcal{F}_{obj}^i \in \mathbb{R}^{N \times C}$, and $N$ and $C$ denote the sequence length and channel dimension, respectively. To transform this feature sequence into a compact condition for object generation, we follow the query-based interaction in~\cite{chang2025reconviagen}: learnable shape queries progressively aggregate the structural and orientation-related information through multi-layer cross-attention,
\begin{equation}
    C_{pose}^i = \text{Attention}(Q_{shape}, \mathcal{F}_{obj}^i, \mathcal{F}_{obj}^i),
\end{equation}
where \(Q_{shape} \in \mathbb{R}^{M \times C}\) denotes the learnable shape queries and \(\mathcal{F}_{obj}^i\) serves as both the key and value. The resulting \(C_{pose}^i\) distills the geometry cues relevant to the current object and serves as the condition for the subsequent rotation-aware generation process. We omit the instance superscript $i$ below for brevity.

\paragraph{Rotation-aware Shape Generation.}
Given the rotation-aware representation \(C_{pose}\) as the conditional input, we aim to synthesize a 3D shape using a DiT-based diffusion model, ensuring that the generated object’s orientation is aligned with the input observation.

To incorporate the 3D spatial cues encoded in \(C_{pose}\), we augment each standard DiT block from~\cite{hunyuan3d2025hunyuan3d} with an additional cross-attention layer, which serves as an effective interface to inject geometric information into the generation process.



We now detail the formulation of the augmented DiT block, i.e., the shape-DiT illustrated in Figure~\ref{fig:pipeline}. For each transformer block indexed by \(k\), let \(h_k\) denote the input latent representation of the \(k\)-th block. We first apply Layer Normalization to \(h_k\), and then perform a cross-attention operation conditioned on \(C_{pose}\) to incorporate the 3D rotation information, expressed as:
\begin{align}
    \hat{h}_k &= h_k + \mathrm{Attention}(\mathrm{LN}(h_k), C_{pose}, C_{pose}).
\end{align}

Subsequently, the output \(\hat{h}_k\) is fed into the standard transformer block, which consists of a layer normalization followed by a feed-forward network. The output of the feed-forward network is then added back to \(\hat{h}_k\) via a residual connection, yielding the input latent for the next block \(h_{k+1}\). This process can be written as:
\begin{align}
    h_{k+1} &= \hat{h}_k + \mathrm{FFN}(\mathrm{LN}(\hat{h}_k)).
\end{align}

Through the above process, the model dynamically attends to relevant orientation cues at each denoising step, guided by the rotation-aware condition \(C_{pose}\), thereby enabling the generation of a rotation-aware 3D mesh. Subsequently, we use a 3D texture generation method to generate a fully textured 3D asset: we generate multi-view images of $\tilde{\mathbf{I}}_o^i$ and extract geometric features from $\mathcal{O}_{\text{pose-aware}}^i$, then feed both into a diffusion model to synthesize texture maps, which are finally unwrapped to UV space to render final textured objects.

\subsection{Scene-Level Reconstruction}
\label{subsec:Position_reconstruction}

After generating the rotation-aware object representations, the scene-level
branch estimates their translation and scale. Unlike joint RTS
estimation~\cite{meng2026scenegen,shi2026scenemaker}, this branch does not treat
rotation as an independent placement output.

We employ the shared geometry encoder \(\mathcal{G}\) (i.e.,
VGGT-\(\Omega\)) to extract both scene- and object-level cues. The full scene
image \(I\) yields \(\mathcal{F}_{scene}\), which captures holistic geometry,
layout, and relative spatial context, while each object-centric crop \(I^i_o\)
yields \(\mathcal{F}_{obj}^i\), which captures local structure and scale cues
for the target object.

For each object \(i\), learnable position queries \(Q_{pos}^i\) retrieve
instance-specific cues from \(\mathcal{F}_{obj}^i\) while incorporating the
global context in \(\mathcal{F}_{scene}\). We formulate the position-aware
attention as:
\begin{equation}
    TS_{pos}^i = \mathrm{Attention}(Q_{pos}^i, \mathcal{F}_{obj}^i, \mathcal{F}_{scene}),
    \label{TS}
\end{equation}
%
where \(TS_{pos}^i\) denotes the learned translation-and-scaling feature token for the \(i\)-th object, which directly encodes the final translation vector \(\mathbf{t}_i \in \mathbb{R}^3\) and isotropic scaling factor \(s_i \in \mathbb{R}\). Leveraging these parameters, we scale and translate each rotation-aware object and place it in the final global scene space, expressed as follows:
\begin{equation}
    \mathcal{O}_{\text{final}}^i = s_i \cdot \mathcal{O}_{\text{pose-aware}}^i + \mathbf{t}_i,
\end{equation}
where \(\mathcal{O}_{\text{pose-aware}}^i\) denotes the rotation-aware object obtained from Sec.~\ref{subsec:pose_generation}.

\subsection{Training Loss}
The overall training objective consists of two terms. For the rotation-aware 3D generation (Sec.~\ref{subsec:pose_generation}), we adopt a flow-matching objective and the training loss is defined as:
\begin{equation}
    \mathcal{L}_{\text{gen}} = \mathbb{E}_{z_0, \epsilon, t, c} \left[ \| v_{\theta}(z_t, t, c) - (z_0 - \epsilon) \|_2^2 \right],
\label{eq:gen_loss}
\end{equation}
where \(z_t = (1-t)\epsilon + t z_0\) defines the linear interpolation between the Gaussian noise \(\epsilon \sim \mathcal{N}(0, \mathbf{I})\) and the ground-truth latent \(z_0\), \(v_{\theta}\) is the predicted velocity field, \(t \in [0,1]\) is the time step, and \(c\) is the input object image.

For the scene-level reconstruction (Sec.~\ref{subsec:Position_reconstruction}), we directly take the predicted \(TS_{pos}^i\) from Eq.~\ref{TS} as the translation and scaling parameters. The loss is then computed as the L1 distance between it and its ground-truth counterpart, expressed as:
\begin{equation}
    \mathcal{L}_{\text{pos}} = \frac{1}{N} \sum_{i=1}^{N} \| TS_{pos}^i - \text{GT}(TS_{pos}^i) \|_1,
\label{eq:pos_loss}
\end{equation}
where \(\text{GT}(TS_{pos}^i)\) is the ground-truth vector for object \(i\).


\begin{figure*}[!t]
    \centering
    \includegraphics[width=0.98\linewidth]{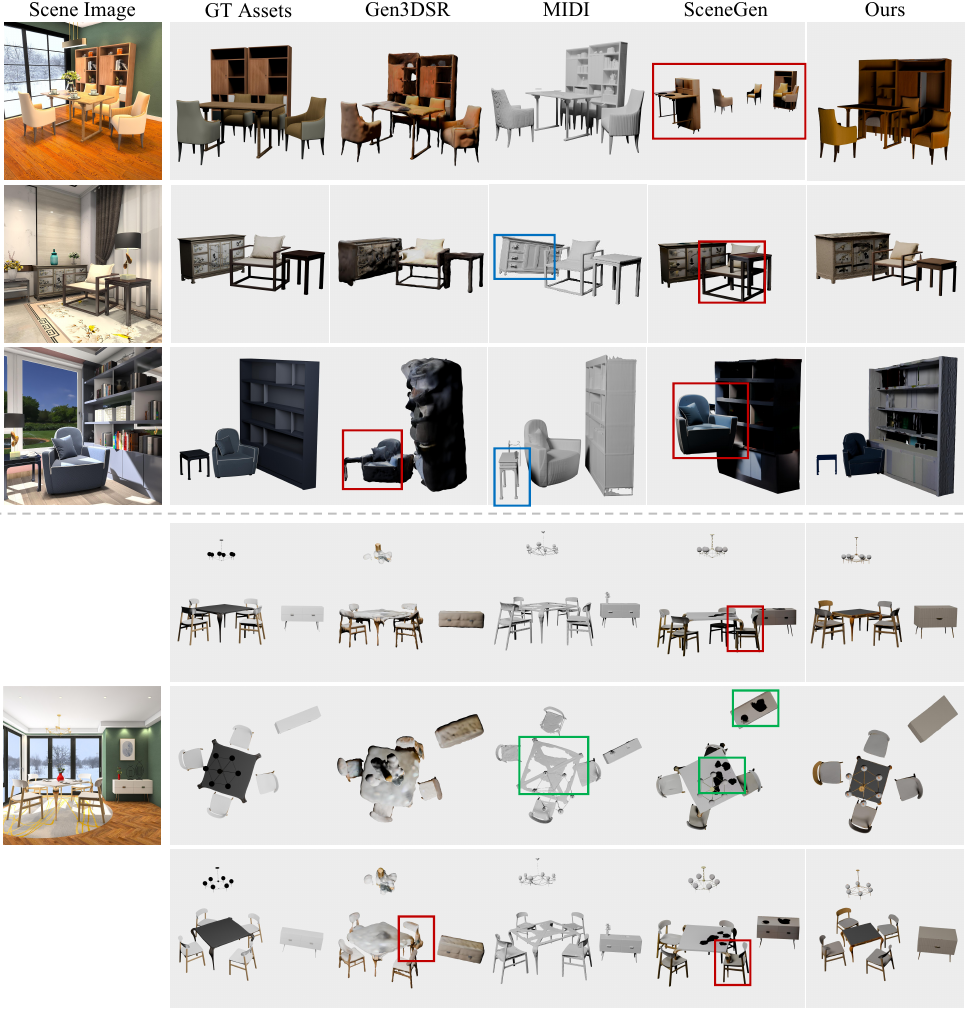}
    \caption{Qualitative comparison across different methods. The panels below the dashed line show results from different viewpoints given the same reference image. Red boxes highlight methods affected by wrong relational placement (e.g., geometric intersections), blue boxes indicate inconsistencies in object orientation, and green boxes reveal incomplete object geometry.}
    \label{fig:experiment_results}
\end{figure*}


\begin{table*}[t]
\centering
\begin{tabularx}{\linewidth}{l *{5}{>{\centering\arraybackslash}X}}
\toprule
\multirow{2}{*}{\textbf{Method}} & \multicolumn{2}{c}{\textbf{Scene-Level}} & \multicolumn{2}{c}{\textbf{Object-Level}} & \multirow{2}{*}{\textbf{$\text{IoU}_B \uparrow$}} \\
\cmidrule(lr){2-3} \cmidrule(lr){4-5}
 & \textbf{CD-S $\downarrow$} & \textbf{F-Score-S $\uparrow$} & \textbf{CD-O $\downarrow$} & \textbf{F-Score-O $\uparrow$} & \\
\midrule
MIDI       & 0.067 & 53.95 & 0.051 & 59.14 & 0.186 \\
Gen3DSR    & 0.055 & 61.18 & 0.095 & 41.69 & 0.227 \\
SceneGen   & \underline{0.017} & \underline{81.49} & \textbf{0.031} & \textbf{71.20} & \underline{0.410} \\
ShapeR & 0.045 & 52.33 & 0.104 & 41.80 & 0.109 \\
SAM 3D     & 0.048 & 66.71 & \underline{0.042} & 68.84 & 0.152 \\
TRELLIS.2  & 0.065 & 62.75 & - & - & - \\
\midrule
\textbf{Ours} & \textbf{0.009} & \textbf{89.50} & \textbf{0.031} & \underline{68.95} & \textbf{0.536} \\
\bottomrule
\end{tabularx}
\caption{Quantitative comparison on the 3D-FUTURE evaluation subset. CD-S and F-Score-S measure scene-level geometry, CD-O and F-Score-O measure object-level geometry, and $\text{IoU}_B$ measures 3D bounding-box alignment. Up and down arrows indicate whether higher or lower values are better, respectively. The best results are in bold, the second-best are \underline{underlined}.}
\label{tab:sota_compare}
\end{table*}

\section{Experiments}
\subsection{Experimental Settings}
We adopt both the rotation-aware diffusion backbone and the texture-generation model from~\cite{hunyuan3d2025hunyuan3d}.
We use the AdamW optimizer with a cosine learning rate scheduler and a maximum learning rate of
$1 \times 10^{-5}$. The model is trained for 500k iterations on 96 Ascend 910B NPUs with a per-device batch size of 4.
To stabilize gradient updates, we apply gradient clipping with a maximum norm of 1.0. For conditional generation, we employ classifier-free guidance (CFG) with a condition dropout probability of 0.1. We use VGGT-\(\Omega\) as our geometry encoder.

\paragraph{Dataset.}
Following standard practices in 3D scene generation, where most prior works \cite{meng2026scenegen, xiang2026native, xia2026points} conduct evaluations on the 3D-FUTURE dataset~\cite{fu20213d}, we utilize this dataset to guarantee fair and rigorous comparisons.
It contains 14,761 training scenes and 5,479 test scenes. Each scene consists of multiple objects, each paired with a rendered image and the corresponding ground-truth segmentation masks.
Given the enormous size of this dataset, we follow the sampling strategy proposed in \cite{ardelean2025gen3dsr}. We randomly sample 150 scenes from 3D-FUTURE to form our test subset for evaluating all competing methods.

\paragraph{Evaluation Metrics.}
Following prior work~\cite{meng2026scenegen}, we uniformly sample point clouds from the generated and ground-truth mesh surfaces. We use FilterReg~\cite{8954154} to align them, as it provides faster and more accurate registration than the traditional Iterative Closest Point (ICP) algorithm.
To evaluate reconstruction quality, we adopt two metrics at both the scene and object levels: Chamfer Distance (CD) measures the bidirectional geometric similarity between the generated and ground-truth point clouds, while F-Score assesses both the accuracy of predicted surface points and the coverage of the ground-truth surface. Additionally, we employ Volumetric Intersection over Union of bounding boxes ($\text{IoU}_B$) to evaluate the accuracy of spatial locations and bounding-box overlaps with respect to the ground-truth scenes.

\paragraph{Evaluation Baselines.}
We compare our method with several state-of-the-art approaches for 3D scene generation and reconstruction. The selected baselines fall into two main categories: per-object optimization methods and full-scene reconstruction methods.
Among the per-object optimization baselines, \textbf{Gen3DSR}~\cite{ardelean2025gen3dsr}, \textbf{SAM 3D}~\cite{sam3dteam2025sam3d3dfyimages}, and \textbf{ShapeR}~\cite{siddiqui2026shaper} first complete the geometry of individual objects via diffusion-based generation and then arrange the reconstructed objects in the scene according to a predefined scene layout or guidance derived from depth and segmentation maps.
Among the full-scene reconstruction methods, \textbf{MIDI}~\cite{huang2025midi} employs a multi-instance attention mechanism to predict all object meshes and their positional parameters. \textbf{SceneGen}~\cite{meng2026scenegen} leverages an end-to-end feed-forward network to generate the geometry, texture, and relative positions of multiple objects simultaneously. \textbf{TRELLIS.2}~\cite{xiang2026native} directly synthesizes all objects in the scene within a field-free sparse voxel structure.

\subsection{Qualitative and Quantitative Comparisons}

\textbf{Qualitative Comparison.} Figure~\ref{fig:experiment_results} presents a qualitative comparison of different methods for reconstructing a 3D scene from a single reference image. We organize the visible differences along three attributes. \textit{Object completeness.} In the shown examples, the MIDI outputs contain visible mesh holes, while parts of the SceneGen reconstructions appear as black regions (see green boxes); SceneReGen retains more complete object surfaces. \textit{Orientation consistency.} Compared with the corresponding ground-truth assets, MIDI reconstructs the table in the second row and the chair in the third row with noticeable orientation discrepancies, as highlighted by the blue boxes. By contrast, SceneReGen more closely matches the ground-truth orientations of both objects in these examples. \textit{Relational placement.} In the first row, SceneGen changes the relative spacing among the furniture compared with the GT assets. Besides, across the lower multi-view panels, SceneReGen better preserves the \textit{relational placement} between the chairs and the table and exhibits fewer visible intersections than the comparison outputs highlighted by red boxes.

\textbf{Quantitative Comparison.} Table~\ref{tab:sota_compare} reports the quantitative results for the evaluated scene generation and reconstruction methods. SceneReGen achieves the best CD-S, F-Score-S, and $\text{IoU}_B$ among the evaluated methods. At the object level, it ties SceneGen for the best CD-O and ranks second on F-Score-O. These results show that the complete SceneReGen system improves scene-level reconstruction and bounding-box alignment while maintaining competitive object-level geometric quality.

\textbf{Cross-Domain Application Potential.} Figure~\ref{fig:teaser} connects object-centric scene reconstruction to two application domains in which explicit 3D object representations are valuable. For autonomous driving, complete, orientation-aware vehicle meshes and their scene-level placement provide an asset-level representation that could complement perception outputs for scenario reconstruction and simulation. For embodied AI, reconstructing a complete manipulable object, such as the bottle shown in Figure~\ref{fig:teaser}, can provide object geometry and pose cues for interaction planning and embodied simulation. Together, these examples illustrate how SceneReGen could connect perception with downstream systems through an asset-centric 3D representation: reconstructed objects remain individually accessible for simulation or interaction, while scene-level placement retains the spatial context needed for reasoning.

\subsection{Ablation Study}
\textbf{Ablation Study on Geometry Encoder and Occlusion Augmentation.}
We conduct ablation experiments to compare the geometry encoders and occlusion augmentation settings in Table~\ref{tab:ablation_stage2}. Variants (a)--(c) compare a generic 2D visual encoder, DINOv2, with two geometry-grounded encoders, VGGT and VGGT-\(\Omega\), while keeping \(M_{\text{occ}}\) disabled. Both geometry-grounded encoders improve all four reconstruction metrics over DINOv2, suggesting that explicitly geometry-grounded representations provide more effective conditioning cues for our reconstruction task than generic 2D visual features. VGGT-\(\Omega\) further improves all four metrics over VGGT and achieves the strongest performance among the evaluated encoders under the same augmentation setting.

Variants (c) and (d) retain VGGT-\(\Omega\) and differ in the use of \(M_{\text{occ}}\). Adding the augmentation reduces CD-S from 0.010 to 0.009 and CD-O from 0.045 to 0.031, while increasing F-Score-S from 87.64 to 89.50 and F-Score-O from 60.52 to 68.95. These consistent improvements support including \(M_{\text{occ}}\) in the final configuration under our evaluation protocol.

\subsection{Limitations}

Our method has the following limitations. (i) Our texture synthesis relies on the native module of Hunyuan3D~\cite{hunyuan3d2025hunyuan3d}, which exhibits limited robustness under challenging visual conditions (e.g., heavy occlusions and extreme lighting), thereby restricting high-fidelity texture generation in these scenarios. (ii) Our approach suffers from performance degradation when provided with low-resolution or blurry scene inputs, mainly due to incomplete reconstruction cues that lead to inaccurate estimation of translation and scale.

\begin{table}[t]
\centering
\setlength{\tabcolsep}{1pt} 
\footnotesize 
\begin{tabular}{l c c c c c c}
\toprule
\multirow{2}{*}{\textbf{Variant}} & \multirow{2}{*}{$\mathcal{G}$} & \multirow{2}{*}{$M_{\text{occ}}$} & \multicolumn{2}{c}{\textbf{Scene-Level}} & \multicolumn{2}{c}{\textbf{Object-Level}} \\
\cmidrule(lr){4-5} \cmidrule(lr){6-7}
& & & \textbf{CD-S $\downarrow$} & \textbf{F-Score-S $\uparrow$} & \textbf{CD-O $\downarrow$} & \textbf{F-Score-O $\uparrow$} \\
\midrule
(a)   & DINOv2        & $\times$ & 0.032 & 75.32 & 0.083 & 49.21 \\
(b)   & VGGT          & $\times$ & 0.012 & 85.21 & 0.058 & 59.31 \\
(c)   & VGGT-$\Omega$ & $\times$ & 0.010 & 87.64 & 0.045 & 60.52 \\
\midrule
(d) \textbf{Ours} & \textbf{VGGT-$\Omega$} & $\checkmark$ & \textbf{0.009} & \textbf{89.50} & \textbf{0.031} & \textbf{68.95} \\
\bottomrule
\end{tabular}
\caption{Ablation study on the geometry encoder $\mathcal{G}$ and the occlusion augmentation strategy $M_{\text{occ}}$. 
$M_{\text{occ}}$ indicates whether our occlusion augmentation is applied during training. 
Best results are in bold.} \label{tab:ablation_stage2}
\end{table}
\section{Conclusion}
We presented SceneReGen, an object-centric framework that turns rotation-aware
object generation into single-image 3D scene reconstruction through selective
pose factorization. Under this factorization, each object's observed orientation
is represented directly in a centered, scale-normalized generated mesh, while
translation and scale are estimated from instance-level and global scene cues.
Geometry-aware shape queries condition a pretrained 3D generator to complete
individual objects, and
scene-aware position queries assemble the generated assets in a shared
observation-aligned scene frame. On the 3D-FUTURE evaluation subset, SceneReGen
achieves the best scene-level CD, scene-level F-Score, and 3D bounding-box IoU
among the evaluated methods, ties the best object-level CD, and ranks second in
object-level F-Score. Together, these results indicate improved complete-scene
reconstruction and spatial alignment while retaining competitive object-level
geometry. The cross-domain outputs further illustrate the potential of this
asset-centric representation for autonomous-driving scenario reconstruction and
embodied-AI interaction.

\bibliography{aaai2027}

\clearpage

\section{Experimental Details}
\subsection{Training Data Preparation}
To endow our rotation-aware generative model with strong generalization across autonomous driving and embodied AI scenarios, we assemble and preprocess a multi-source dataset consisting of Objaverse, 3D-FUTURE and MeshFleet. Specifically, 3D-FUTURE supplies high-quality indoor furniture assets; MeshFleet provides diverse 3D vehicle models for driving scenes; Objaverse covers commonplace daily commodities such as bottles and cups, which are essential for embodied interactive environments.

To guarantee overall dataset quality, we first render the objects into multiple views to filter out unqualified instances—including assets with degraded visual appearance, oversimplified geometric structures, and clustered multi-object messy meshes. Afterwards, we conduct a watertightness inspection over all mesh samples. Non-watertight surfaces severely hinder the training convergence of diffusion VAEs; Therefore, we eliminate flawed samples by measuring the Chamfer Distance between raw meshes and their VAE reconstructions, discarding instances with excessive reconstruction error.

The proposed multi-stage filtering pipeline yields a refined high-quality dataset containing around 25K object meshes. Following standard protocols from prior work \cite{xiang2026native}, we render 24 multi-view RGB images for each object with randomly sampled azimuth and elevation angles. Critically, we align ground-truth meshes with the camera viewpoints of paired input images to maintain consistency with our rotation-aware generative optimization objective.

\subsection{Evaluation Details}
We adopt dataset-specific evaluation standards across all test benchmarks, with elaborate configuration details illustrated as follows. Consistent with previous literature \cite{meng2026scenegen}, geometric quality is measured within a normalized canonical 3D space bounded by \(x,y,z\in[-1,1]\). Before metric calculation, rigid point-cloud registration is performed to align each generated mesh with its corresponding ground truth. For the 3D-FUTURE benchmark \cite{fu20213d}, all quantitative metrics are computed under the dataset’s native geometric scale. We uniformly sample 10,000 points from both generated meshes and ground-truth meshes, and the distance threshold for F-Score calculation is fixed at 0.1.

\subsection{Inference Details}
During inference, we crop individual object regions based on provided instance masks and apply center padding to resize all input images to a fixed resolution of \(512 \times 512\). For conditional generation, we set the classifier-free guidance (CFG) dropout probability to 0.1. All experiments share identical inference hyperparameters: we employ 50 denoising sampling steps with a CFG scale of 3.0 to output the final latent representations for mesh generation.

\section{More Qualitative Results}
This section presents additional qualitative visualizations on the test split of the 3D-FUTURE dataset \cite{fu20213d}. For each furniture instance reconstructed by our method, we render five distinct viewing angles to fully display the complete rotation-aware generated mesh produced by SceneReGen.

\begin{figure*}[!t]
    \centering
    \includegraphics[width=0.98\linewidth]{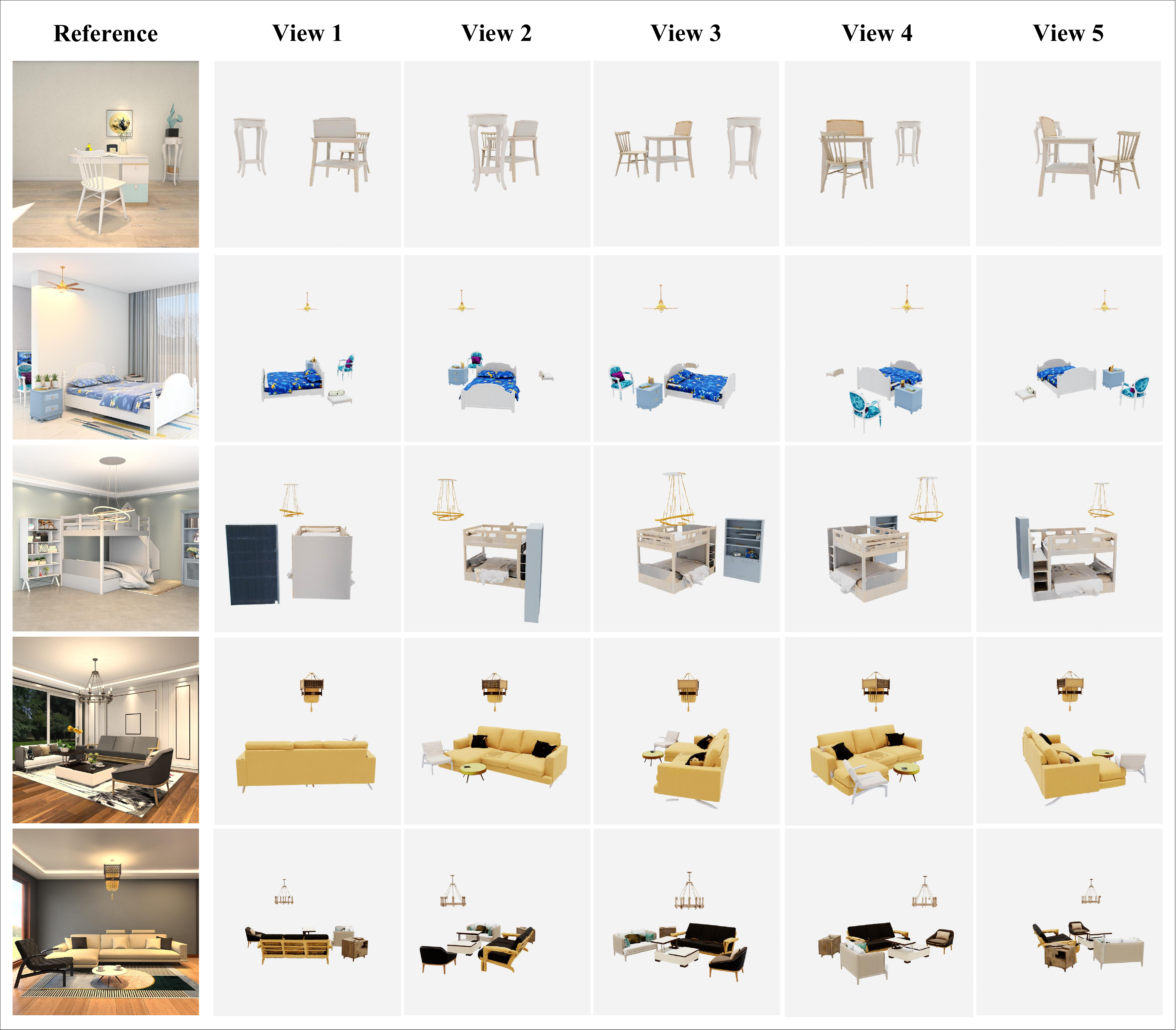}
    \caption{More qualitative results from our method.}
    \label{fig:mor_experiment_results}
\end{figure*}

In Figure~\ref{fig:mor_experiment_results}, these comprehensive multi-view comparisons intuitively validate that our method maintains stable geometric integrity and orientation consistency under arbitrary camera viewpoints.

\section{Limitations and Future Work}
\subsection{Limitations}
Our SceneReGen framework has three main limitations:
First, directly using the off-the-shelf Hunyuan3D texture module lacks occlusion-aware optimization, leading to blurriness and color distortion on heavily occluded objects.
Second, reconstruction performance degrades significantly with low-resolution or blurry input images due to the loss of fine-grained geometric and structural cues.
Third, the translation-and-scale branch predicts placement independently without inter-object collision regularization, frequently resulting in mesh penetration and unnatural overlaps in dense indoor scenes.

\subsection{Future Work}
We outline three directions for future research:
First, we will design an occlusion-adaptive texture refinement branch utilizing auxiliary occlusion masks features to improve texture quality.
Second, we plan to train on larger and more diverse multi-domain 3D scene datasets to boost generalization and alleviate data distribution bias.
Third, we will integrate explicit physical priors and collision-aware regularization into the position estimation module to prevent mesh penetration and ensure natural spatial layouts.

\end{document}